\documentclass[pdflatex,sn-mathphys-num]{sn-jnl}

\usepackage{graphicx}%
\usepackage{multirow}%
\usepackage{amsmath,amssymb,amsfonts}%
\usepackage{amsthm}%
\usepackage{mathrsfs}%
\usepackage[title]{appendix}%
\usepackage{xcolor}%
\usepackage{array}%
\usepackage{textcomp}%
\usepackage{manyfoot}%
\usepackage{booktabs}%
\usepackage{algorithm}%
\usepackage{algorithmicx}%
\usepackage{algpseudocode}%
\usepackage{listings}%
\usepackage{hyperref}
\usepackage{lmodern}
\usepackage{bookmark}
\usepackage{lipsum}
\usepackage{anyfontsize}

\begin{document}

\title[Lit-Net]{Harnessing Multi-resolution and Multi-scale Attention for Underwater Image Restoration}


\author*{\fnm{Alik} \sur{Pramanick}}\email{p.alik@iitg.ac.in}

\author{\fnm{Arijit} \sur{Sur}}\email{arijit@iitg.ac.in}

\author{\fnm{V. Vijaya} \sur{Saradhi}}\email{saradhi@iitg.ac.in}

\affil{\orgdiv{Department of Computer Science and Engineering}, \orgname{Indian Institute of Technology Guwahati}, \country{India}}


\abstract{
Underwater imagery is often compromised by factors such as color distortion and low contrast, posing challenges for high-level vision tasks. Recent underwater image restoration (UIR) methods either analyze the input image at full resolution, resulting in spatial richness but contextual weakness, or progressively from high to low resolution, yielding reliable semantic information but reduced spatial accuracy. Here, we propose a lightweight multi-stage network called Lit-Net that focuses on multi-resolution and multi-scale image analysis for restoring underwater images while retaining original resolution during the first stage, refining features in the second, and focusing on reconstruction in the final stage. Our novel encoder block utilizes parallel $1\times1$ convolution layers to capture local information and speed up operations. Further, we incorporate a modified weighted color channel-specific $l_1$ loss ($cl_1$) function to recover color and detail information. Extensive experimentations on publicly available datasets suggest our model's superiority over recent state-of-the-art methods, with significant improvement in qualitative and quantitative measures, such as $29.477$ dB PSNR ($1.92\%$ improvement) and $0.851$ SSIM ($2.87\%$ improvement) on the EUVP dataset. The contributions of Lit-Net offer a more robust approach to underwater image enhancement and super-resolution, which is of considerable importance for underwater autonomous vehicles and surveillance. The code is available at: \url{https://github.com/Alik033/Lit-Net}.}




\maketitle

\section{Introduction}\label{secI}
\begin{figure}[ht]
\centering
\includegraphics[width=3.2in,height=2in,clip]{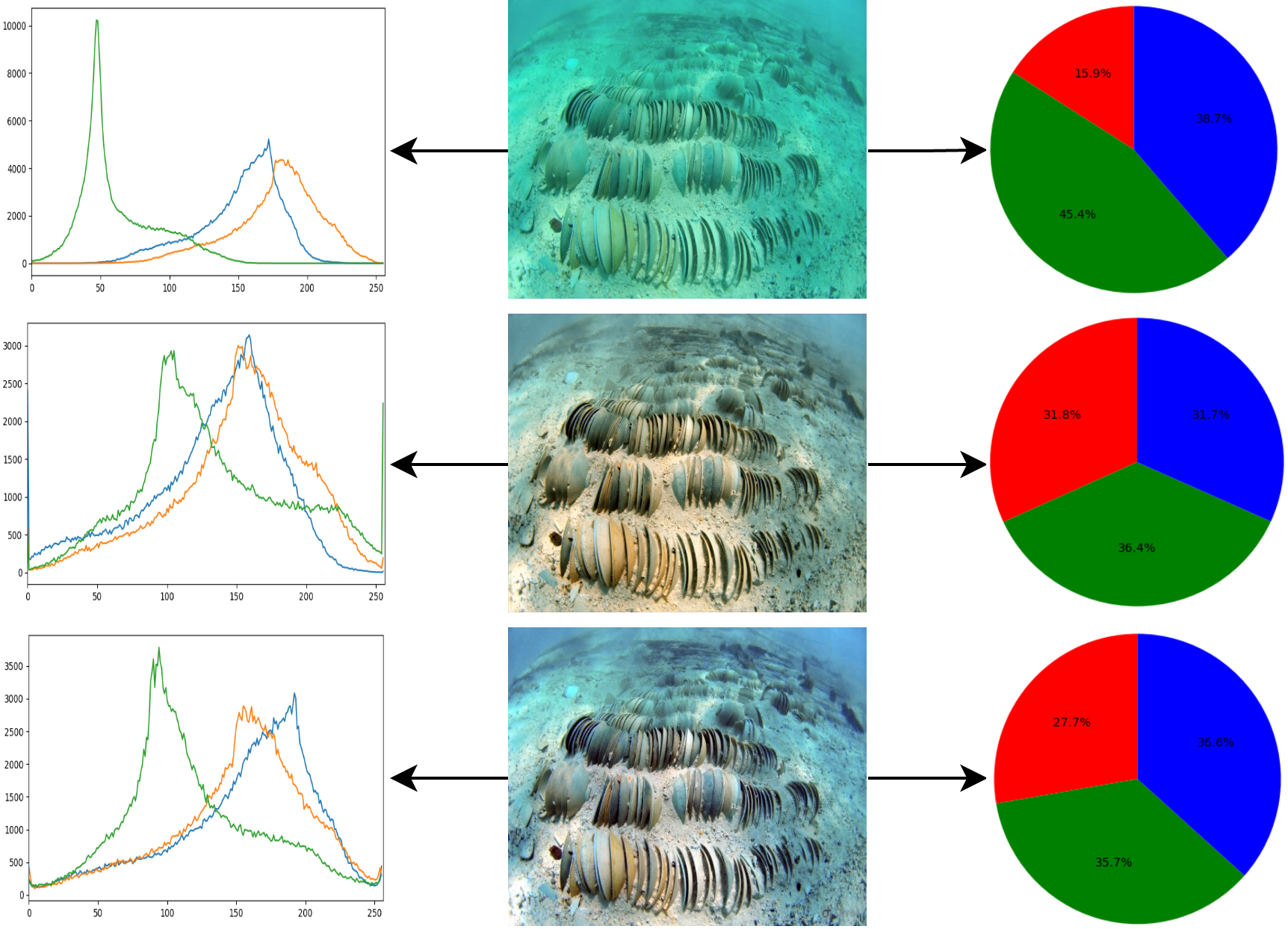}
\caption{First row represents the tricolor histogram and pie chart of the degraded underwater image,\textbf{ second row represents the tricolor histogram and pie chart of the enhanced underwater image by our Lit-Net}, and third row represent the tricolor histogram and pie chart of the ground-truth image.}
\label{dist}
\end{figure}
In the recent past, underwater vision has become an emerging research topic. Autonomous underwater vehicles (AUVs) need to understand their operating environment in various applications, such as monitoring marine life and reefs, examining submarine cables and wreckage, and collaborating with humans. The above-mentioned applications are performed by executing high-level vision tasks on underwater images and videos, such as object classification~\cite{rokh2023comprehensive}, detection~\cite{xu2023systematic}, segmentation~\cite{yi2024coordinate}, etc. Visual degradation may have an adverse effect on these high-level vision tasks.

Even though advanced cameras are commonly used in underwater photography and videography, improved clarity is still needed due to the limited visibility caused by light bending, absorption, and scattering, which can cause distortions in the captured images and videos. One of the primary reasons for the visual distortions in underwater imaging is the non-uniform attenuation of light, which varies depending on the wavelength of the light \cite{chiang2011underwater}. Further, in the underwater scenario, blue color traverses longer than other colors as its wavelength is the shortest. The prevalence of blue and green colors in underwater imagery directly impacts the effectiveness of higher-level tasks \cite{raveendran2021underwater}. Fig. \ref{dist} demonstrates the tricolor histogram and pie chart color distribution of a degraded underwater image, an enhanced version of the same image, and the ground truth image, respectively. There is a noticeable issue with color casting in underwater images. As a result, underwater images are often of low contrast, blurry, bluish, greenish, etc. In order to tackle these issues, it is necessary to employ efficient low-level computer vision techniques such as image enhancement, super-resolution, and other similar methods.

In recent research, deep learning-based methods have exhibited remarkable outcomes when applied to low level vision tasks. It has been noticed that a lot of State-of-the-art Underwater Image Restoration (UIR) works process the R-G-B channels with the same receptive field sizes. However, the same receptive fields for each R-G-B channel may not be suitable in most underwater situations. Sharma et al. \cite{sharma2023wavelength} have shown that different sizes of receptive fields help in UIR. Still, it needs to capture the desired global effects that require considering the relationships between color components. Deep neural networks designed for UIR are broadly categorized into the single scale (high-resolution) \cite{sharma2023wavelength,li2019underwater, islam2020simultaneous} or an encoder-decoder feature processing \cite{fabbri2018enhancing, chen2023mngnas,islam2020fast}. High-resolution (single-scale) networks avoid down-sampling, resulting in images with more spatially precise information. However, these networks become less beneficial at encoding semantic information. For encoder-decoder based approaches, the encoder gradually maps the input to a low-resolution representation, and the decoder reverses the mapping to obtain the original resolution. These strategies grasp a broad context by reducing spatial resolution; however, finer spatial details are lost, making image reconstruction harder. 

Underwater image restoration is a technique that requires pixel-to-pixel correspondence between the input and output images and is sensitive to the position of pixels. Knowing that the underwater images are mainly bluish and greenish, removing only the undesired degraded content is crucial while preserving all the essential spatial information, including true edges and texture. For this, different sizes of receptive fields are used for different color channels (such as R ($3 \times 3$), G ($5 \times 5$), and B ($7 \times 7$)) in the proposed architecture. This helps us achieve multi-resolution analysis for different color channels to maintain spatial (high-resolution) information and capture the local and global context. Additionally, we process the whole RGB image with $1\times1$ convolution to capture the desired global effects considering the relationships between color components. However, a 1×1 convolution outputs an image with the same width and height as the input. It acts as a channel collapse as well as a color messenger to the preceding layers \cite{szegedy2015going}. A new multi-scale strategy is incorporated to maintain semantic (encoder-decoder) information. Apart from that, we use the 1x1 convolution layer in our encoder network to speed up the number of operations. Furthermore, we have incorporated attention-based skip connections for more discriminative representations.
\begin{figure}[ht]
  \centering
  \begin{tabular}{@{}c@{}}
    \includegraphics[width=10cm, height=0.9in]{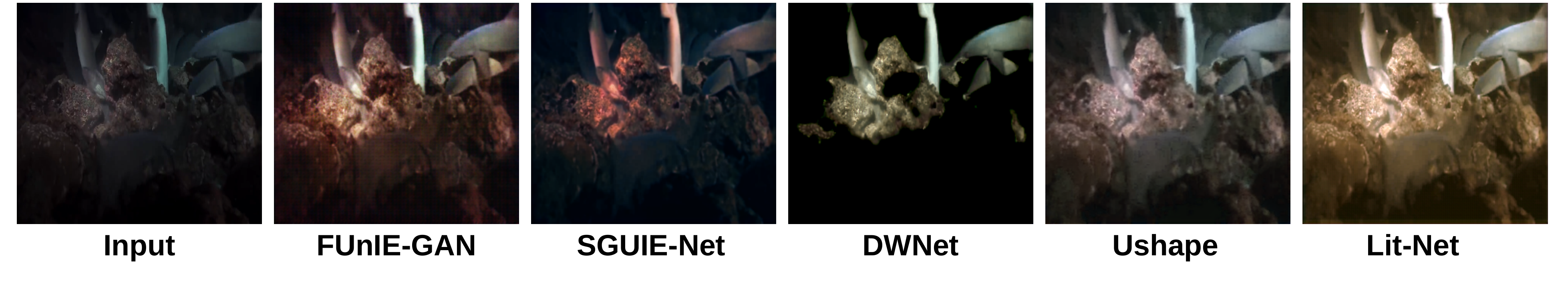} \\[\abovecaptionskip]
    \small (a)
  \end{tabular}
  \begin{tabular}{@{}c@{}}
    \includegraphics[width=10cm, height=0.9in]{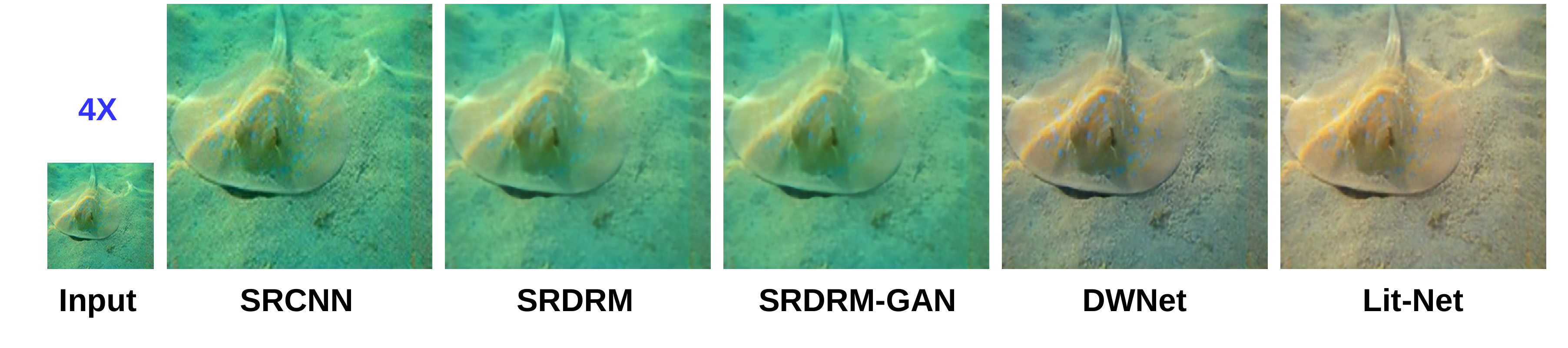} \\[\abovecaptionskip]
    \small (b)
  \end{tabular}
  \caption{Visual comparisons of (a) enhancement example on low contrast underwater image, (b) 4x super-resolve example on underwater image. The proposed Lit-Net model enhanced the contrast and removes the color deviation.}\label{low_con}
\end{figure}

Fig. \ref{low_con}(a) visually compare the proposed scheme and recent best-published works on the low-contrast real underwater image. It is observed that the proposed Lit-Net method produced visually more accepted results than recent best-published works. Fig. \ref{low_con}(b) illustrates similar comparison results for the underwater image super-resolution (UISR) method. It is observed that most of the existing methods generate a greenish tone in their resulting image, while Lit-Net yields a visually improved color-corrected result. The major contributions of this work is outlined as follows:
\begin{itemize}
    \item We propose a lightweight multi-stage network for underwater image enhancement and super-resolution. The first stage is used for multi-resolution image analysis while retaining its original resolution. The second stage enhances the intermediate features while retraining the important information. The last stage focus on the reconstruction of the enhanced and super-resolved image.
    \item We design a novel encoder block where each encoder layer has four parallel 1x1 convolution layers to pay attention to the dominant color and capture the local information precisely. It also speeds up the number of operations as it is less expensive than the larger-size of filters.
    \item We modified the $l_{1}$ loss function to a weighted color channel specific $l_{1}$ loss ($cl_{1}$) function. It helps to recover the color and detail information.
\end{itemize}
We provide extensive experiments against the best-published state-of-the-art in underwater image enhancement (UIE) and UISR to show the benefit. The performance of the proposed model has been demonstrated by comparing it with existing methods through various tasks, including underwater semantic segmentation and object detection.
\section{Related Work}\label{secRW}
In this section, we briefly describe the overview of prior research on UIR. The two main parts of this literature are (1) underwater image enhancement and (2) underwater single-image super-resolution.
\subsection{\textbf{Underwater Image Enhancement }}Earlier, approaches are separated as physical model-based or physical model-free types. Physical model-based techniques \cite{drews2016underwater,galdran2015automatic,li2016underwater,
peng2018generalization,peng2017underwater} typically use an imaging model and estimate the attenuation light and transmission map with hand-crafted priors to restore underwater images. The commonly used priors are the dark channel prior (DCP) \cite{drews2013transmission,he2010single,wen2013single} and its modified versions red channel prior \cite{li2016single}, blurriness prior \cite{peng2017underwater}, light attenuation prior \cite{song2018rapid}, transmission prior~\cite{ouyang2024multi}. Derya et al. \cite{akkaynak2019sea} aims to retrieve color with the revised model using RGBD images. However, statistical priors may perform poorly in difficult underwater situations, and the optical image formation model may yield errors due to irregular scene effects \cite{akkaynak2018revised,akkaynak2017space}. 

Physical model-free strategies include multi-scale fusion that combines information from different levels of resolution \cite{ancuti2017color,ancuti2011fusion,aguirre2022evaluation}, variational optimization that involves the optimization of a cost function with respect to a set of parameters \cite{fu2014retinex}, the work proposed in pixel distribution adjustment \cite{ghani2015underwater,xue2023underwater}, focus on altering image intensity values to make them more realistic. The absence of the underwater imaging model in physical model-free techniques can create unwanted visual anomalies due to the interference of noise in the underwater environment.

In recent times, deep learning-based approaches for underwater image enhancement have achieved substantial gains \cite{qiao2023underwater,li2019underwater,fabbri2018enhancing,islam2020fast,li2021underwater,li2020underwater,li2017watergan,
uplavikar2019all,wu2021two,tolie2024dicam,tao2024multi,lai2023two}. Previous studies have utilized Generative Adversarial Networks (GANs) to address the issue of inadequate underwater images and their corresponding clean image pairs. By using GANs, they synthesize underwater image datasets or execute unpaired training. In \cite{guo2019underwater}, they used GAN \cite{goodfellow2014generative} to generate the large synthetic dataset and proposed a two-stage architecture for color correction of monocular underwater images. Liu et al. \cite{liu2019underwater} used the Cycle-GAN \cite{engin2018cycle} to produce roughly 4000 pairs of synthetic underwater degraded and ground truth images. Also, they proposed UResnet, which would use residual learning to improve underwater images. Their UResnet used the VDSR \cite{kim2016accurate} model (presented for the super-resolution model). \cite{uplavikar2019all} presented a straightforward classifier to produce the GAN model better discriminative for various water types. The authors of \cite{wang2020gan} proposed a conditional GAN approach that relies on the attention \cite{vaswani2017attention,lin2021eapt} mechanism for the activity of UIR. The work proposed in \cite{han2020underwater} suggested the use of a spiral Generative Adversarial Network based network for UIE. Li et al. \cite{li2019underwater} created the Underwater Image Enhancement Benchmark (UIEB) dataset, which is a comprehensive dataset of real-world underwater images for the purpose of UIE. Further, the work in \cite{li2019underwater} proposed Water-Net, a gate fusion CNN-based model for UIE by fusing three enhanced inputs: (a) gamma-corrected input, which affects the overall brightness and contrast of the image, (b) white-balanced input, which ensure that the colors in an image appear accurate and natural, regardless of the lighting conditions under which the image was taken, and (c) histogram equalized input. Islam et al. \cite{islam2020fast} presented the EUVP dataset containing 20K underwater images. Also, \cite{islam2020fast} introduced a conditional GAN-based fully-convolutional framework for UIE in real time. To address color cast and low contrast degradation concerns \cite{li2021underwater} proposed a multi-color space encoder framework that incorporates an attention mechanism to merge the attributes of various color spaces into a cohesive structure and adaptively choose the most significant features. Recently, Huo et al. \cite{huo2021efficient} presented a deep learning model that progressively refines underwater images using a wavelet boost learning technique in spatial and frequency domains. Qi et al. \cite{qi2022sguie} proposed  SGUIE-Net for UIE, in which they introduced semantic information as high-level guidance.
\subsection{\textbf{Underwater Image Super-resolution}} The objective of Single Image Super-Resolution (SISR) is to improve the resolution of a low-quality image by generating a higher resolution version of the same image \cite{jin2019flexible,tian2020coarse,yang2019deep,zhang2018residual}. Li et al. \cite{li2019underwater} proposed SRCNN, the initial Super-Resolution (SR) model utilized a Color Features-based methodology and Multi-Layer Perceptron (MLP) technique. The work in \cite{chen2020wavelet} proposed an SR network based on wavelet transform that decomposes an image into a series of wavelets that vary in scale and frequency and improve dense block structure. Lu et al. \cite{lu2017underwater} initially offered a self-similarity SR algorithm to the LR images for HR images. Finally, they got the SR images by applying a convex fusion rule that combines multiple estimates or observations into a single estimate of the HR images. The authors of \cite{islam2020underwater} build a multi-modal optimization technique for underwater super-resolution that quantifies the perceptual quality of the image based on factors such as global content, color, and local style information to create an adversarial training pipeline. Islam et al. \cite{islam2020simultaneous}  presented a generative framework based on residual-in-residual networks that simultaneously improve and super-resolving underwater imagery for better visual perception. Sharma et al. \cite{sharma2023wavelength} presented a deep learning-based framework using CNN, which consists of multiple stages, and it employs the distinct size of the kernel to each R-G-B color channel of an image, guided by its wavelength for concurrent UIE and UISR task. Recently, Zang et al. \cite{zhang2022attention} proposed a multi-path network guided by an attention module that mainly utilizes the cross-convolution technique for UISR.
\subsection{\textbf{Major Observation}}
Despite significant advancements in the enhancement of underwater images, current approaches still suffer from visual artifacts, such as color distortion, poor visibility, low contrast, and hazy. These are not particularly efficient in improving images that are severely degraded and lacking in texture. In such instances, the resulting images may be over-saturated due to the amplification of noise. While the correction of hue is typically accurate, there is still a deficiency in the restoration of both texture and color. Within a GAN-based model \cite{islam2020fast,zhang2023underwater}, the discriminator can become overly proficient too quickly, leading to a decrease in the gradient effect and ultimately inhibiting the generator's ability to learn. In such instances, the produced images often exhibit poor color uniformity and inadequate texture details. In \cite{qi2022sguie}, they use a segmentation mask with a degraded image to produce the enhanced result. Generating the segmentation mask itself will take time, and if we use a pre-trained segmentation module, it won't produce good segmentation for a diverse set of underwater images, which will affect the UIE performance. In the case of UISR, even though efforts have been made to super-resolve the images, they often contain unwanted artifacts, such as a lack of representation of texture features and high-frequency information. Due to variations in attenuation ranges across channels, underwater images require special considerations that differ from those necessary for outdoor images. As a result, directly applying outdoor models \cite{dudhane2020end,dharejo2021twist,jiang2022photohelper} may not be appropriate in underwater settings. Inspired by \cite{sharma2023wavelength}, we have shown how the different receptive field sizes help to gain performance in both UIE and UISR. Also, we have shown how the lightweight encoder-decoder based model with attention-based skipped connection is sufficient to generate notable performance gain in both tasks. Taking into account structural appropriateness and texture similarity, the proposed approach can generate high-quality results without the need for any post-processing operations. 
\begin{figure*}[ht]
\centering
\includegraphics[width=\textwidth]{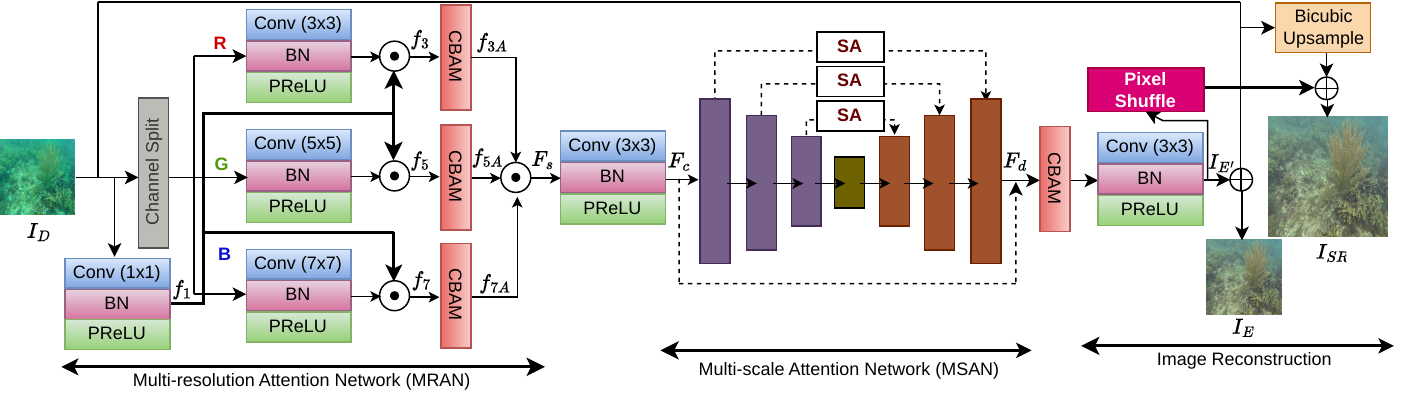}
\caption{Overview of our proposed model for UIE and UISR. The model accepts a degraded underwater image as input and produces an improved image that is visually and spatially enhanced.}
\label{fig2}
\end{figure*}
\section{Proposed Method}\label{secPM}
\subsection{\textbf{Network Architecture}}
In this work, a deep learning-based architecture is proposed to enhance and super-resolve underwater images. The overall architecture is depicted in Fig. \ref{fig2}. We divided the network into three distinct modules as follows:
\begin{itemize}
    \item Multi-resolution Attention Network (MRAN)
    \item Multi-scale Attention Network (MSAN)
    \item Image Reconstruction
\end{itemize}
The Multi-resolution Attention Network (MRAN) extracts shallow features that correspond to low-frequency details from the input degraded underwater image ($I_{D}$). The purpose of MRAN is to analyze the input image in a multi-resolution way while maintaining its original resolution. To obtain the most prominent features, followed by the MRAN module, we design a multi-scale attention network (MSAN) by extracting the rich features related to high-frequency details from the input image while retraining the important information through attention-based skipped connection. Finally, an image reconstruction module is devised to restore the enhanced and super-resolved image.
\subsubsection{\textbf{Multi-resolution Attention Network}} In the proposed multi-resolution attention network (MRAN), we directly use full-resolution input images to characterize multi-resolution feature representations using different kernel sizes (having different receptive fields) for analyzing the global and local information. This module incorporates four convolutional layers using four different kernel sizes of $1 \times 1$, $3\times3$, $5\times5$, and $7\times7$ with batch normalization and PReLU activation. We generate the R, G, and B channel-specific features with $3\times3$, $5\times5$, and $7\times7$ kernels. Also, we extract the features from the whole RGB image with $1\times1$ kernel. Then, we concatenate RGB features with channel-specific features. The concatenated features are passed through attention blocks and finally concatenated to get the output of the MRAN block. Shallow contextual features ($F_{s}$) are acquired given the input $I_{D}$.
\begin{equation}
    F_{s} = f_{3A} \odot f_{5A} \odot f_{7A},
\end{equation}
where, $\odot$ indicates the concatenation operation and the attentive contextual features ($f_{kA}$) with receptive field size $k$ can be estimated as
\begin{equation} 
    \begin{split}
        f_{3A} = A_{n}(f_{3}) \\
        f_{5A} = A_{n}(f_{5}) \\
        f_{7A} = A_{n}(f_{7})
    \end{split}
\end{equation}
\begin{figure}[ht]
\centering
\includegraphics[width=3.5in,height=2in]{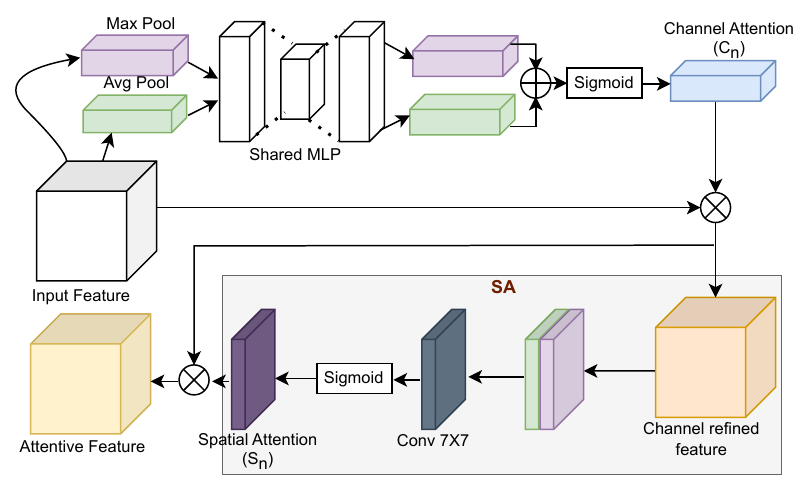}
\caption{Network architecture of CBAM block \cite{woo2018cbam}.}
\label{fig_cbam}
\end{figure}

We use CBAM \cite{woo2018cbam} module as an attention block ($A_{n}$) to refine the obtained features $f_{k}$. The architecture of CBAM is shown in Fig. \ref{fig_cbam}. CBAM inputs an intermediate feature map to retrieve the channel and spatial attention features. The purpose of channel attention is to focus on identifying the important features present in an input image. It aims to determine 'what' aspects of the image are significant in terms of representing the desired information. Spatial attention concentrates on 'where' is an informative part. We recommend that readers refer to \cite{woo2018cbam} for more information on CBAM. The refine features $f_{k}$ can be computed as    
\begin{equation} 
    \begin{split}
    	f_{1} = p_{r}(b_{n}(c^{1\times 1}(I_{D})))\\
        f_{3} = p_{r}(b_{n}(c^{3\times 3}(I_{D}^{R}))) \odot f_{1} \\
        f_{5} = p_{r}(b_{n}(c^{5\times 5}(I_{D}^{G}))) \odot f_{1} \\
        f_{7} = p_{r}(b_{n}(c^{7\times 7}(I_{D}^{B}))) \odot f_{1}
    \end{split}
\end{equation}
where $c^{k \times k}$, $b_{n}$, and $p_{r}$ represent a convolution operation, batch normalization \cite{ioffe2015batch} and parametric ReLU layers, respectively. $k\times k$ indicates the kernel size. $I_{D}^{C}$ ($C \in R, G, B$) indicates the individual channel of a degraded image.

The output generated by MRAN is first fed into a convolution layer, which is then used as input for MSAN. 
\begin{equation}
    F_{c} = p_{r}(b_{n}(c^{3 \times 3}(F_{s})))
\end{equation}
\subsubsection{\textbf{Multi-scale Attention Network}} 
\begin{figure}[ht]
\centering
\includegraphics[width=12cm,height=1.8in]{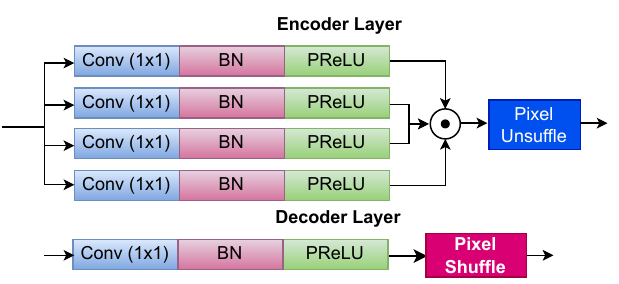}
\caption{Network architecture of proposed encoder and decoder layer.}
\label{fig4}
\end{figure}

The proposed MSAN module is a UNet-like architecture where each encoder and corresponding decoder layer is linked with an attention-based skipped connection. In this work, a novel extension (refer to Fig. \ref{fig4}) is used in each encoder layer to analyze the different input resolutions more accurately. To do this, we map the high-resolution feature $(F_{c})$ to a low-resolution representation and slowly retrieve the high-resolution representation from the lower level. Generally, the encoder-decoder process learns semantically-richer feature representation $(f_{ds})$ (multi-scale contextual information), but it is not spatially precise due to down-sampling. To get both semantically and spatially richer features $(F_{d})$, the features map $f_{ds}$ is concatenated with the high-resolution features map $(F_{c})$. Also, to ensure the feature re-usability, we use spatial attention-based skip connection, as in Fig. \ref{fig2}. 
The output of MSAN can be estimated as
\begin{equation}
    F_{d} = F_{c} \odot f_{ds}
\end{equation}

We have modified each encoder layer (refer to Fig. \ref{fig4}). Generally, a normal encoder layer uses a $3\times3$ convolutional layer and a maxpool layer which is computationally costly because of $3\times3$ kernel. Here, each encoder layer contains four $1\times1$ convolution layers in parallel. Finally, we concatenated the four convoluted features and passed them through a pixel unshuffle layer. Instead of a maxpool layer, we use pixel-unshuffle to downsample each feature to minimize the information loss. We use three encoder layers in our MSAN. Mathematically, the encoder is formulated as
\begin{equation} 
    \begin{split}
        E_{1}^{1} = p_{r}(b_{n}(c^{1\times 1}(F_{c}))) \\
        E_{1} = P_{unshuffle}(E_{1}^{1} \odot E_{1}^{1} \odot E_{1}^{1} \odot E_{1}^{1})\\
        E_{2}^{1} = p_{r}(b_{n}(c^{1\times 1}(E_{1})))\\
        E_{2} = P_{unshuffle}(E_{2}^{1} \odot E_{2}^{1} \odot E_{2}^{1} \odot E_{2}^{1})\\
        E_{3}^{1} = p_{r}(b_{n}(c^{1\times 1}(E_{2})))\\
        E_{3} = P_{unshuffle}(E_{3}^{1} \odot E_{3}^{1} \odot E_{3}^{1} \odot E_{3}^{1})
    \end{split}
\end{equation}
where $E_{i}^{1}$ represents the each branch of each encoder layer $E_{i}$ ($i=1,2,3.$) and $P_{unshuffle}$ indicates the pixel-unshuffle operation.
The output of the encoder can be computed as
\begin{equation}
    B_{kn} = p_{r}(b_{n}(c^{1\times 1}(E_{3})))
\end{equation}
Let $S_{n}$ represent the spatial attention module and $P_{shuffle}$ represents the pixel shuffle operation as an up-sampling operation to increase the spatial resolution. The decoder can be defined as      
\begin{equation} 
    \begin{split}
        D_{1} = P_{shuffle}(B_{kn}) \odot S_{n}(E_{3})\\
        D_{2} = P_{shuffle}(D_{1}) \odot S_{n}(E_{2}) \\
        f_{ds}=D_{3} = P_{shuffle}(D_{2}) \odot S_{n}(E_{1})
    \end{split}
\end{equation}
where, $D_{i}$ ($i = 1 , 2 , 3.$) indicates the each decoder layer.
\subsubsection{\textbf{Image Reconstruction}} The final stage of the proposed model is used as the reconstruction module. In the case of underwater image enhancement, it takes the output features of MSAN $(F_{d})$ as input and feds them into a convolutional layer to produce residual image $(I_{E^{'}})$. Finally, the enhanced image is obtained by $I_{E} = (I_{E^{'}}) + I_{D}$, where
\begin{equation}
    I_{E^{'}} = p_{r}(b_{n}(c^{3 \times 3}(F_{d})))
    \label{9}
\end{equation}
We use three channels to get enhanced underwater images.
In the case of underwater image super-resolution, we use $3s^{2}$ ($s$ indicates the scale factor) channel for equation \ref{9} as it upsampled the MSAN features to the desired resolution. After that, we use pixel-shuffle \cite{shi2016real} layers to refine the spatial quality of the image further. The purpose of this layer is to enhance the resolution and details of the image. Also, we use bicubic up-sampling to improve the spatial resolution of the $I_{D}$. Finally, the super-resolved image ($I_{SR}$) is obtained by
\begin{equation}
    I_{SR} = P_{shuffle}(I_{E}) + U_{bicubic}(I_{D})
\end{equation}
\subsection{\textbf{Loss Function}}
\subsubsection{\textbf{Weighted Color channel specific L1 loss}}
It is observed in the literature that the $l_{2}$ loss function can lead to the appearance of blurry artifacts in the resulting enhanced images. Due to this, we have chosen to employ $l_{1}$ loss rather than $l_{2}$ loss. To train Lit-Net, instead of using simple $l_{1}$ loss, we have used weighted color channel-wise $l_{1}$ loss for reconstructing high-quality underwater images. This is because underwater images can be affected differently by the water medium in each channel, and thus, the $cl_{1}$ loss enables the model to focus more on the channels that are dominated in underwater. The weighted channel-specific $l_{1}$ loss can be expressed as follows: 
\begin{equation}
    cl_{1} = w_{r} l_{1_{r}} + w_{g} l_{1_{g}} + w_{b} l_{1_{b}}
\end{equation}
where, $l_{1_{r}}$, $l_{1_{g}}$, and $l_{1_{b}}$ indicates the red (r), green (g), and blue (b) channel specific $l_{1}$ loss, these can be formulated as
\begin{equation}
\begin{split}
     l_{1_{r}} =\frac{1}{n} \sum_{i=1}^{n} || ES(I_{D_{i}})_{r} - (I_{O_{i}})_{r} ||_{1} \\
     l_{1_{g}} =\frac{1}{n} \sum_{i=1}^{n} || ES(I_{D_{i}})_{g} - (I_{O_{i}})_{g} ||_{1} \\
     l_{1_{b}} =\frac{1}{n} \sum_{i=1}^{n} || ES(I_{D_{i}})_{b} - (I_{O_{i}})_{b} ||_{1}
    \end{split}
\end{equation}
where ES(.) represents the process of Lit-Net, $I_{O}$ is the original clean image in the case of UIE, and the higher-resolution image in the case of UISR. Empirically, we have set $w_{r}$, $w_{g}$, and $w_{b}$ to be equal to 1, 1.5, and 2, respectively.
\subsubsection{\textbf{Perceptual loss}}
Although training with pixel loss enhances PSNR, it does not improve perceptual quality. To address this flaw, we integrated the Perceptual loss \cite{johnson2016perceptual} to make it more perceptually improved. This loss function helps to ensure that the enhanced image maintains its fine details and textures. To accomplish this goal, we have used pre-trained VGG16 $(V_{f}(\phi))$ \cite{simonyan2014very} model, which is pre-trained on a large-scale image database (ImageNet). We define the cost function as the Euclidean distance between the relu2$\_$2 features obtained from the predicted image and the actual clean image, and it can be expressed as follows:
\begin{equation}
    l_{p} = \frac{1}{n} \sum_{i=1}^{n} ||V_{f}(ES(I_{D_{i}});\phi) - V_{f}(I_{O_{i}};\phi)||_{2}^{2}
\end{equation}
\subsubsection{\textbf{SSIM loss}}
Besides using $l_{1}$ and perceptual losses, to minimize the structural dissimilarities between $I_{E}$ and $I_{O}$, we have included the SSIM loss \cite{wang2004image} to our approach. The SSIM is a way to quantify how similar two images are in terms of their structure, luminance, and contrast and can be expressed as:
\begin{equation}
    SSIM(c) = \frac{2 . \mu_{a} . \mu_{b}+K_{1}}{\mu_{a}^{2}+\mu_{b}^{2}+K_{1}} . \frac{2 . \sigma_{ab}+K_{2}}{\sigma_{a}^{2}+\sigma_{b}^{2}+K_{2}}
\end{equation}
where a,b are patches from $I_{E}$ or $I_{SR}$, and $I_{O}$, respectively, $\mu$, $\sigma$, and $\sigma_{ab}$ denote mean, standard deviation and covariance, given $c$ the center of patches. $K_{1}$ and $K_{2}$ are fixed parameters. The SSIM loss that we have included in our method can be expressed as follows:
\begin{equation}
    l_{s} = \frac{1}{2n} \sum_{i=1}^{n} 1-SSIM(ES(I_{D_{i}}), I_{O_{i}})
\end{equation}
 Finally, our proposed method is trained by minimizing the following losses:
\begin{equation}
    l_{T} = \lambda_{1}cl_{1} + \lambda_{p} l_{p} + \lambda_{s} l_{s}
\end{equation}
where $\lambda_{1}$, $\lambda_{p}$, and $\lambda_{s}$ have been experimentally set to 1, 0.02, and 0.5, respectively.
\section{Experiments}\label{secE}
In this section, we perform experiments on four popular standard datasets to show the efficacy of our Lit-Net model. Three of them are for underwater image enhancement tasks, namely, EUVP \cite{islam2020fast}, UIEB \cite{li2019underwater}, and SUIM-E \cite{qi2022sguie}. The remaining dataset is for underwater single image super-resolution task, namely, UFO-120 \cite{islam2020simultaneous}. We evaluate the performance of our Lit-Net model against state-of-the-art approaches and conduct an ablation study to assess the impact of each component of Lit-Net. 
\begin{table*}[!ht]
  \caption{MSE, PSNR, SSIM, MS-SSIM, LPIPS, UIQM, UISM, and BRISQUE comparison on EUVP dataset with the best-published works for UIE. First, second, and third best performance are represented in \textcolor{red}{red}, \textcolor{blue}{blue}, and \textcolor{green}{green} colors, respectively. $\downarrow$ indicates lower is better.}
  \label{tab:euvp}
  \centering
  \small
  \setlength{\tabcolsep}{1pt}
  \begin{tabular}{l|c|c|c|c|c|c|c|c}
    \hline    Method&MSE$\downarrow$&PSNR&SSIM&UIQM&LPIPS$\downarrow$&UISM&MS-SSIM&BRISQUE$\downarrow$\\
    \hline \hline
    UGAN \cite{fabbri2018enhancing}&0.355&$26.551$&\textcolor{green}{$0.807$}&$2.896$&0.220&6.833&0.936&35.859\\
    UGAN-P \cite{fabbri2018enhancing}&0.347&26.549&0.805&2.931&0.223&6.816&0.935&35.099\\
    FUnIE-GAN\cite{islam2020fast}&0.386&26.220&0.792&2.971&0.212&\textcolor{green}{6.892}&0.924&\textcolor{blue}{30.912}\\
    FUnIE-GAN-UP&0.600&25.224&0.788&2.935&0.246&6.853&0.925&34.070\\
    Deep SESR \cite{islam2020simultaneous} &\textcolor{green}{0.325}&\textcolor{green}{27.081}&0.803&\textcolor{red}{3.099}&\textcolor{black}{0.206}&\textcolor{red}{7.051}&\textcolor{green}{0.940}&35.179\\
    DWNet\cite{sharma2023wavelength}&\textcolor{blue}{0.276}&\textcolor{blue}{$28.654$}&\textcolor{blue}{$0.835$}&\textcolor{green}{$3.042$}&\textcolor{blue}{0.173}&\textcolor{red}{7.051}&\textcolor{blue}{0.950}&\textcolor{red}{30.856}\\
    Ushape~\cite{peng2023u}&0.370&26.822&0.811&\textcolor{blue}{3.052}&\textcolor{green}{0.187}&6.843&\textcolor{green}{0.949}&35.648\\
    \hline
    Ours&\textcolor{red}{0.225}&\textcolor{red}{29.477}&\textcolor{red}{0.851}&\textcolor{black}{3.027}&\textcolor{red}{0.169}&\textcolor{blue}{7.011}&\textcolor{red}{0.954}&\textcolor{green}{32.109}\\
  \hline
\end{tabular}
\end{table*}
\subsection{\textbf{Datasets}}
\subsubsection{\textbf{EUVP}} The EUVP dataset contains 11345 image pairs for underwater enhancement tasks. The testing set consists of 515 paired images. The size of each image is 256$\times$256. 
\subsubsection{\textbf{UIEB}} We have used the UIEB dataset for our experiments. It contains 890 image pairs. Following \cite{li2019underwater}, randomly, 800 images choose to train the model, and the remaining 90 images are kept aside for testing the model. We have resized the image into 256$\times$256 for training and testing. Also, we use the 60 challenge images of the UIEB dataset for testing the model without ground truth references.
\subsubsection{\textbf{SUIM-E}} The SUIM-E dataset contains a total of 1635 images, out of which 1525 images were used for training and the remaining 110 images were reserved for testing. We make the image size $256\times256$ for training and testing.

Apart from the datasets mentioned above, we have also used the underwater color cast set (\textbf{UCCS})~\cite{liu2020real} and \textbf{U45}~\cite{li2019fusion} dataset for testing.
\subsubsection{\textbf{UFO-120}} We employed the UFO-120 dataset for performing underwater image super-resolution task. It contains 1620 image pairs. We utilized 1500 images for the purpose of training and 120 images for testing. In this dataset, degraded LR images of the dimension 320$\times$240 and the corresponding HR images of the dimension $640\times480$ are present for 2$\times$ super-resolution. To perform 3$\times$ and 4$\times$ super-resolution, we have down-sample the degraded LR images.
\subsection{\textbf{Experimental Setup}}
\subsubsection{\textbf{Training Process}}
The proposed Lit-Net model is implemented based on an open-source deep learning framework called Pytorch. We have performed two independent training for UIE and UISR task. In the training phase, we utilized the Adam optimizer with an initial learning rate of $2e-4$. All the experiments have been conducted on a Ubuntu 20.04 LTS operating system and one Nvidia A100 GPU. During the training process, a batch size of 5 was employed for the model.
\subsubsection{\textbf{Evaluation Metric}}
To ensure a fair comparison between the proposed model and existing state-of-the-art approaches, we have assessed the performance of the model using widely used reference and reference-less image quality metrics. These include Mean-squared Error (MSE), Peak Signal-to-Noise Ratio (PSNR), Structural Similarity Index Measure (SSIM), Multi-scale Structural Similarity Index Measure (MS-SSIM), Learned Perceptual Image Patch Similarity (LPIPS), Underwater Image Quality Measure (UIQM) \cite{panetta2015human}, Underwater Image Sharpness Measure (UISM), and Blind/Reference-less Image Spatial Quality Evaluator (BRISQUE).
\subsubsection{\textbf{State-of-the-Art Methods}}
\begin{itemize}
\item \textbf{Underwater Image Enhancement:} In the objective of UIE, the proposed work is compared with several existing state-of-the-art studies, which include: UDCP \cite{drews2013transmission} $(ICCVW-13)$, GBdehaze \cite{li2016single} $(ICASSP-16)$, IBLA \cite{peng2017underwater} $(TIP-17)$, ULAP \cite{song2018rapid} $(PCM-18)$, CBF \cite{ancuti2017color} $(TIP-18)$, GDCP~\cite{peng2018generalization} $(TIP-18)$ UGAN \cite{fabbri2018enhancing} $ (ICRA-18)$, WaterNet~\cite{li2019underwater} $(TIP-20)$, UWCNN~\cite{li2020underwater} $(PR-20)$, FUnIE GAN \cite{islam2020fast} $(RAL-20)$, DeepSESR \cite{islam2020simultaneous} $(RSS-20)$, UIEC\^{} 2~\cite{wang2021uiec} $(SPIC-21)$, Ucolor \cite{li2021underwater} $TIP-21$, UWCNN-SD~\cite{wu2021two} $(IJOE-21)$ MILE~\cite{zhang2022underwater} $(TIP-22)$, PUIE~\cite{fu2022uncertainty} $(ECCV-22)$, SGUIE-Net \cite{qi2022sguie} $(TIP-22)$, UGIF-Net~\cite{zhou2023ugif} $(TGRS-23)$, DWNet~\cite{sharma2023wavelength} $(TOMM-23)$ and Ushape~\cite{peng2023u} $(TIP-23)$. 
\item \textbf{Underwater Image Super-resolution:} In the task of UISR, we compare the proposed work against the following best-published methods: SRCNN \cite{dong2015image} $(TPAMI-16)$, SRResNet \cite{ledig2017photo} $(CVPR-17)$, SRGAN \cite{ledig2017photo} $(CVPR-17)$, SRDRM \cite{islam2020underwater} $(ICRA-20)$, Deep SESR \cite{islam2020simultaneous} $(RSS-20)$, and DWNet\cite{sharma2023wavelength} $(TOMM-23)$.
\end{itemize}
Our proposed method employs identical training and test sets as other deep learning-based approaches. For a fair comparison, the training and testing image resolution was the same for all the methods as ours. Apart from that, we use the same environment (Python/Matlab) for calculating the measurement metrics.
\begin{table*}[ht]
  \caption{PSNR, SSIM, MS-SSIM, LPIPS, UIQM, UISM, and BRISQUE comparison on UIEB test set with the best-published works for UIE. First, second, and third best performances are represented in \textcolor{red}{red}, \textcolor{blue}{blue}, and \textcolor{green}{green} colors, respectively. $\downarrow$ indicates lower is better.}
  \label{tab:uieb}
  \centering
  \small
  \setlength{\tabcolsep}{3pt}
  \begin{tabular}{l|c|c|c|c|c|c|c}
  \hline
    Method&PSNR&SSIM&MS-SSIM&LPIPS$\downarrow$&UIQM&UISM&BRISQUE$\downarrow$\\
    \hline \hline
    UDCP \cite{drews2013transmission}&13.026&0.545&0.769&0.283&1.922&\textcolor{green}{7.424}&24.133\\
    GBdehaze \cite{li2016single}&15.378&0.671&0.777&0.309&2.520&\textcolor{red}{7.444}&23.929\\
    IBLA \cite{peng2017underwater}&19.316&0.690&0.855&0.233&2.108&\textcolor{blue}{7.427}&\textcolor{green}{23.710}\\
    ULAP \cite{song2018rapid}&19.863&0.724&0.865&0.256&2.328&7.362&25.113\\
    CBF \cite{ancuti2017color}&20.771&0.836&0.890&0.189&\textcolor{green}{3.318}&7.380&\textcolor{blue}{20.534}\\
    \hline
    UGAN \cite{fabbri2018enhancing}&23.322&0.815&\textcolor{green}{0.932}&0.199&\textcolor{red}{3.432}&7.241&27.011\\
    UGAN-P \cite{fabbri2018enhancing}&\textcolor{blue}{23.550}&0.814&\textcolor{blue}{0.933}&0.192&\textcolor{blue}{3.396}&7.262&25.382\\
    FUnIE-GAN\cite{islam2020fast}&21.043&0.785&0.890&0.173&3.250&7.202&\textcolor{black}{24.522}\\
    SGUIE-Net \cite{qi2022sguie}&\textcolor{green}{23.496}&\textcolor{blue}{0.853}&0.926&\textcolor{blue}{0.136}&3.004&7.362&24.607\\
    DWNet \cite{sharma2023wavelength} &23.165&\textcolor{green}{0.843}&0.929&\textcolor{green}{0.162}&2.897&7.089&24.863\\
    Ushape \cite{peng2023u}& 21.084&0.744 &0.895&0.220 &3.161&7.183&24.128\\
    \hline
    Ours&\textcolor{red}{23.603}&\textcolor{red}{0.863}&\textcolor{red}{0.935}&\textcolor{red}{0.130}&3.145&7.396&\textcolor{blue}{23.038}\\
    \hline
\end{tabular}
\end{table*}
\subsection{\textbf{Comparisons with State-of-the-Arts}}
Both quantitative and quantitative  comparison results of the proposed model with the recent best-published methods are presented in the following subsections:
\begin{table*}[ht]
  \caption{PSNR, SSIM, MS-SSIM, LPIPS, UIQM, UISM, and BRISQUE comparison on SUIM-E test set with the best-published works for UIE. First, second, and third best performances are represented in \textcolor{red}{red}, \textcolor{blue}{blue}, and \textcolor{green}{green} colors, respectively. $\downarrow$ indicates lower is better}
  \label{tab:suim_e}
  \centering
  \small
  \setlength{\tabcolsep}{3pt}
  \begin{tabular}{l|c|c|c|c|c|c|c}
  \hline
    Method&PSNR&SSIM&MS-SSIM&LPIPS$\downarrow$&UIQM&UISM&BRISQUE$\downarrow$\\
    \hline \hline
    UDCP \cite{drews2013transmission}&12.074&0.513&0.742&0.270&1.648&\textcolor{red}{7.537}&22.788\\
    GBdehaze \cite{li2016single}&14.339&0.599&0.743&0.355&2.255&\textcolor{green}{7.400}&\textcolor{green}{20.175}\\
    IBLA \cite{peng2017underwater}&18.024&0.685&0.849&0.209&1.826&7.341&20.957\\
    ULAP \cite{song2018rapid}&19.148&0.744&0.871&0.231&2.115&\textcolor{blue}{7.475}&21.250\\
    CBF \cite{ancuti2017color}&20.395&0.834&0.884&0.194&\textcolor{red}{3.003}&7.360&21.115\\
    \hline
    UGAN \cite{fabbri2018enhancing}&24.704&0.826&0.941&0.190&2.894&7.175&20.288\\
    UGAN-P \cite{fabbri2018enhancing}&\textcolor{green}{25.050}&0.827&\textcolor{green}{0.943}&0.188&\textcolor{green}{2.901}&7.184&\textcolor{red}{18.768}\\
    FUnIE-GAN\cite{islam2020fast}&23.590&0.825&0.913&0.189&\textcolor{blue}{2.918}&7.121&22.560\\
    SGUIE-Net \cite{qi2022sguie}&\textcolor{red}{25.987}&\textcolor{green}{0.857}&\textcolor{blue}{0.945}&\textcolor{green}{0.153}&2.637&7.090&25.927\\
    DWNet \cite{sharma2023wavelength} &24.850&\textcolor{blue}{0.861}&0.941&\textcolor{blue}{0.133}&2.707&7.381&20.757\\
    Ushape \cite{peng2023u}& 22.647&0.783 &0.917&0.213 &2.873&7.061&22.876\\
    \hline
    Ours&\textcolor{blue}{25.117}&\textcolor{red}{0.884}&\textcolor{red}{0.950}&\textcolor{red}{0.118}&\textcolor{blue}{2.918}&7.368&\textcolor{blue}{19.602}\\
    \hline
\end{tabular}
\end{table*}
\begin{table*}[ht]
  \caption{UISM, UIQM, and BRISQUE comparison on UIEB challenge set with the best-published works. First, second, and third best performances are represented in \textcolor{red}{red}, \textcolor{blue}{blue}, and \textcolor{green}{green} colors, respectively.}
  \label{tab:uieb_challeng}
  \centering
  \setlength{\tabcolsep}{1pt}
  \tiny
  \begin{tabular}{l|c|c|c|c|c|c|c|c|c|c|c|c}
    \hline
    &UDCP &GBdehaze &IBLA &ULAP &CBF &UGAN &FUnIE-GAN &Ucolor &SGUIE-Net &DWNet&Ushape &Ours\\
    &\cite{drews2013transmission}&\cite{li2016single}&\cite{peng2017underwater}&\cite{song2018rapid}&\cite{ancuti2017color}&\cite{fabbri2018enhancing}&\cite{islam2020fast}&\cite{li2021underwater}&\cite{qi2022sguie}&\cite{sharma2023wavelength}&\cite{peng2023u}\\
    \hline \hline
    UISM&7.358&\textcolor{blue}{7.399}&7.310&7.239&\textcolor{green}{7.396}&7.158&7.140&7.244&7.269&6.677&7.292&\textcolor{red}{7.458}\\
    UIQM&1.566&2.280&2.142&1.815&\textcolor{red}{2.810}&2.662&\textcolor{black}{2.768}&2.483&2.527&2.269&\textcolor{green}{2.783}&\textcolor{blue}{2.795}\\
     BRISQUE$\downarrow$&29.658&26.264&\textcolor{black}{24.972}&30.472&29.213&25.118&\textcolor{green}{24.773}&25.093&27.320&31.160&\textcolor{red}{23.616}&\textcolor{blue}{24.755}\\
     \hline
\end{tabular}
\end{table*}
\subsubsection{\textbf{Quantitative Evaluation}}
This sub-section has focused on presenting a quantitative comparison between the proposed method and other existing state-of-the-art approaches with respect to reference and no-reference quality evaluations. The results of these evaluations have been provided in Tables \ref{tab:euvp}, \ref{tab:uieb}, \ref{tab:suim_e}, \ref{tab:uieb_challeng}, \ref{tab:uccs_u45} and \ref{tab:ufosr}. Table \ref{tab:euvp}, \ref{tab:uieb}, \ref{tab:suim_e}, \ref{tab:uieb_challeng}, and \ref{tab:uccs_u45} provide a quantitative comparison of UIE methods. Table \ref{tab:ufosr} presents the evaluation conducted on the task of UISR. Table \ref{tab:euvp} illustrates that the proposed approach has achieved better results than previous methods for full reference-based quality metrics on the EUVP test set. Besides, Lit-Net also achieves the second and third-best results in terms of non-reference measurement. It is observed that the proposed model has gained improvement on both SSIM and PSNR, $1.92\%$ and $2.87\%$, respectively. Lit-Net has a low LPIPS score, indicating that the image patches are perceptually more similar to the ground truth. The Lit-Net has surpassed the performance of previous state-of-the-art methods in terms of PSNR, SSIM, and MS-SSIM on the UIEB test dataset, as demonstrated in Table \ref{tab:uieb}. Additionally, the results also indicate that the proposed model is competitive with the recent schemes in terms of LPIPS and BRISQUE. Table \ref{tab:suim_e} shows that our Lit-Net achieves $2.67\%$ SSIM gains over the previous best methods on the SUIM-E dataset. Our method provides a substantial gain of $0.53\%$ MS-SSIM compared to SGUIE-Net. The corresponding LPIPS score is less, which indicates better perceptual quality. Table \ref{tab:uieb_challeng} demonstrates the UISM, UIQM, and BRISQUE scores of different methods on the UIEB challenge set. The table indicates that the proposed Lit-Net achieves the highest UISM value, second highest BRISQUE value and UIQM value over the state-of-the-art methods. Table \ref{tab:uccs_u45} demonstrates the quantitative results of U45 and UCCS datasets. The proposed approach attains the top UIQM scores for the UCCS and U45 datasets, respectively. Lit-Net exhibits superior performance on various datasets, implying the model's robustness. Table \ref{tab:para_flops} shows that our model takes second less number of parameters and GFLOPs. However, this marginal increment also provides a relative performance boost of up to 2.87\% and 1.92\% regarding PSNR and SSIM over the state-of-the-art on EUVP dataset, which justifies the trade-off between performance and parameters.
\begin{table*}[ht]
  \caption{UIQM comparison on UCCS and U45 datasets with the best-published works. First, second, and third best performances are represented in \textcolor{red}{red}, \textcolor{blue}{blue}, and \textcolor{green}{green} colors, respectively.}
  \label{tab:uccs_u45}
  \centering
  \setlength{\tabcolsep}{1pt}
  \tiny
  \begin{tabular}{l|c|c|c|c|c|c|c|c|c|c|c|c}
    \hline
    &GDCP &WaterNet &FUnIE-GAN &UWCNN &UIEC\^{} 2 &Ucolor &UWCNN-SD&MLIE &PUIE &UGIF-Net &DWNet &Ours\\
    &\cite{peng2018generalization}&\cite{li2019underwater}&\cite{islam2020fast}&\cite{li2020underwater}&\cite{wang2021uiec}&\cite{li2021underwater}&\cite{wu2021two}&\cite{zhang2022underwater}&\cite{fu2022uncertainty}&\cite{zhou2023ugif}&\cite{sharma2023wavelength}\\
    \hline \hline
    UCCS&2.716&3.073&3.074&2.981&2.900&2.983&\textcolor{blue}{3.130}&2.863&2.943&\textcolor{green}{3.112}&3.089&\textcolor{red}{3.167}\\
     \hline
     U45&2.370&3.006&2.563&3.110&3.152&\textcolor{green}{3.185}&3.134&2.550&2.625&\textcolor{blue}{3.264}&3.128&\textcolor{red}{3.279}\\
     \hline
\end{tabular}
\end{table*}
\begin{table*}[ht]
  \caption{Comparison with the model parameters and GFLOPs of the SOTA model with the image size $256 \times 256$. Lower is better.}
  \label{tab:para_flops}
  \centering
  \small
  \setlength{\tabcolsep}{1pt}
  \begin{tabular}{c|c|c|c|c|c|c|c|c}
    \hline
    &WaterNet& UGAN &FUnIE-GAN &Ucolor&SGUIE-Net &DWNet &Ushape &Ours\\
    \hline \hline
    Parameters (M)&24.8&57.17&7.71&157.4&18.55&\textcolor{red}{0.48} & 65.6& \textcolor{blue}{0.54}\\
     \hline
     FLOPs (G)&193.7&18.3&\textcolor{red}{10.7}&443.9&123.5&18.2&66.2&\textcolor{blue}{17.8}\\
     \hline
\end{tabular}
\end{table*}

Table \ref{tab:ufosr} presents the quantitative experiments for single image underwater image super-resolution. It can be inferred from Table \ref{tab:ufosr} that the Lit-Net model has demonstrated superior performance compared to other methods for the task of UISR. We evaluate the performance of our approach against state-of-the-art UISR methods on the UFO-120 test set. Our Lit-Net model obtains the highest PSNR and SSIM scores on a scale of 4, as presented in Table \ref{tab:ufosr}, suggesting that our results closely resemble the ground truth regarding visual features. Our method also produces a comparable UIQM result, displaying underwater aspects.
\begin{center}
\begin{table*}[ht]
  \caption{PSNR, SSIM, and UIQM comparison on UFO-120 dataset with the best-published works for UISR. First, second, and third best performances are represented in \textcolor{red}{red}, \textcolor{blue}{blue}, and \textcolor{green}{green} colors, respectively.}
  \label{tab:ufosr}
  \centering
  \setlength{\tabcolsep}{1.5pt}
  \tiny
  \begin{tabular}{l|c|c|c|c|c|c|c|c|c}
    \hline
             Method& \multicolumn{3}{c|}{PSNR}&\multicolumn{3}{c|}{SSIM}&\multicolumn{3}{c}{UIQM}\\
             \hline
     & 2x & 3x & 4x & 2x & 3x & 4x & 2x & 3x & 4x\\
    \hline
    \hline
    SRCNN \cite{dong2015image}&$24.75 \pm 3.7$&$22.22 \pm 3.9$&$19.05 \pm 2.3$&$.72 \pm .07$&$.65 \pm .09$&$.56 \pm .02$&$2.39 \pm .35$&$2.24 \pm .17$&$2.02 \pm .47$\\
    SRResNet\cite{ledig2017photo}&$25.23 \pm 4.1$&$23.85 \pm 2.8$&$19.13 \pm 2.4$&$.74 \pm .08$&$.68 \pm .07$&$.56 \pm .05$&$2.42 \pm .37$&$2.18 \pm .26$&$2.09 \pm .30$\\
    SRGAN \cite{ledig2017photo}&\textcolor{red}{$26.11 \pm 3.9$}&$23.87 \pm 4.2$&$21.08 \pm 2.3$&$.75 \pm .06$&$.70 \pm .05$&$.58 \pm .09$&$2.44 \pm .28$&$2.39 \pm .25$&$2.26 \pm .17$\\
    SRDRM \cite{islam2020underwater}&$24.62 \pm 2.8$&-&$23.15 \pm 2.9$&$.72 \pm .17$&-&\textcolor{green}{$.67 \pm .19$}&$2.59 \pm .64$&-&$2.56 \pm .63$\\
    SRDRM-GAN\cite{islam2020underwater}&$24.61 \pm 2.8$&-&$23.26 \pm 2.8$&$.72 \pm .17$&-&\textcolor{green}{$.67 \pm .19$}&$2.59 \pm .64$&-&\textcolor{green}{$2.57 \pm .63$}\\
    Deep SESR \cite{islam2020simultaneous} &$25.70 \pm 3.2$&\textcolor{red}{$26.86 \pm 4.1$}&\textcolor{green}{$24.75 \pm 2.8$}&\textcolor{green}{$.78 \pm .08$}&\textcolor{green}{$.75 \pm .06$}&$.66 \pm .05$&\textcolor{red}{$3.15 \pm .48$}&\textcolor{green}{$2.87 \pm .39$}&$2.55 \pm .35$\\
    DWNet \cite{sharma2023wavelength}&\textcolor{green}{$25.71 \pm 3.0$}&\textcolor{green}{$25.23 \pm 2.7$}&\textcolor{blue}{$25.08 \pm 2.9$}&\textcolor{blue}{$.80 \pm .08$}&\textcolor{blue}{$.79 \pm .10$}&\textcolor{blue}{$.74 \pm .14$}&\textcolor{green}{$2.99 \pm .57$}&\textcolor{blue}{$2.96 \pm .60$}&\textcolor{red}{$2.97 \pm .59$}\\
    \hline
    Ours&\textcolor{blue}{$25.78 \pm 2.8$}&\textcolor{blue}{$25.30 \pm 2.8$}&\textcolor{red}{$25.32 \pm 2.8$}&\textcolor{red}{$.82 \pm .08$}&\textcolor{red}{$.80 \pm .09$}&\textcolor{red}{$.80 \pm .09$}&\textcolor{blue}{$3.00 \pm .60$}&\textcolor{red}{$2.99 \pm .57$}&\textcolor{blue}{$2.93 \pm .58$}\\
    \hline
  \end{tabular}
\end{table*}
\end{center}
\subsubsection{\textbf{Qualitative Evaluation}}
\begin{figure}[ht]
\centering
\includegraphics[width=\textwidth]{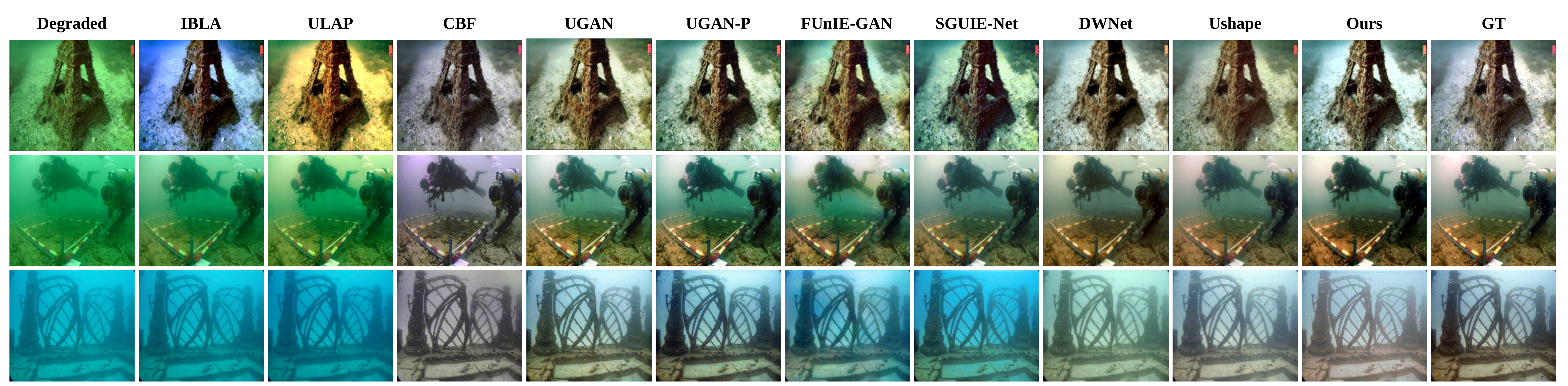}
\caption{Visual demonstration against the state-of-the-art UIE models on UIEB dataset.}
\label{uieb}
\end{figure}

\begin{figure}[ht]
\centering
\includegraphics[width=\textwidth]{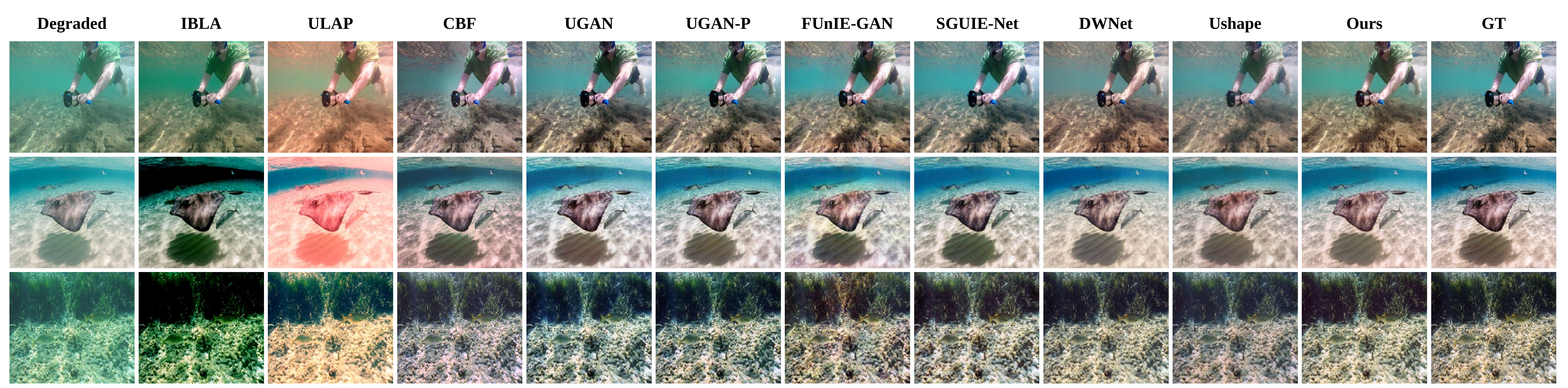}
\caption{Visual demonstration against the state-of-the-art UIE models on SUIM-E dataset.}
\label{suim_e}
\end{figure}

\begin{figure}[ht]
\centering
\includegraphics[width=\textwidth]{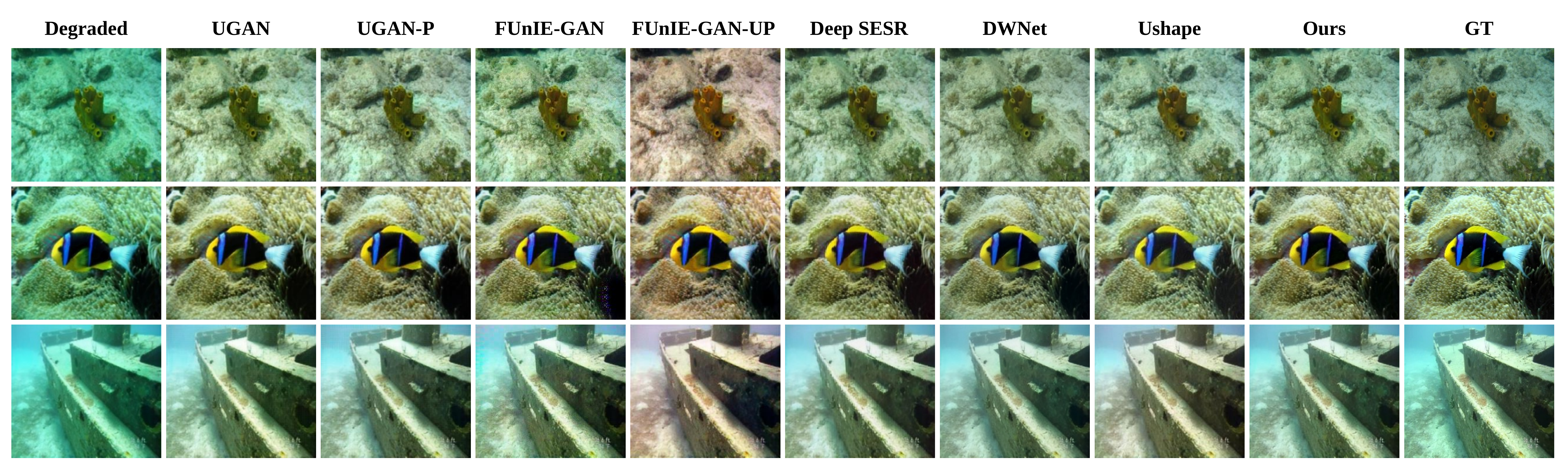}
\caption{Visual demonstration against the state-of-the-art UIE models on EUVP dataset.}
\label{euvp}
\end{figure}

\begin{figure}[ht]
\centering
\includegraphics[width=\textwidth]{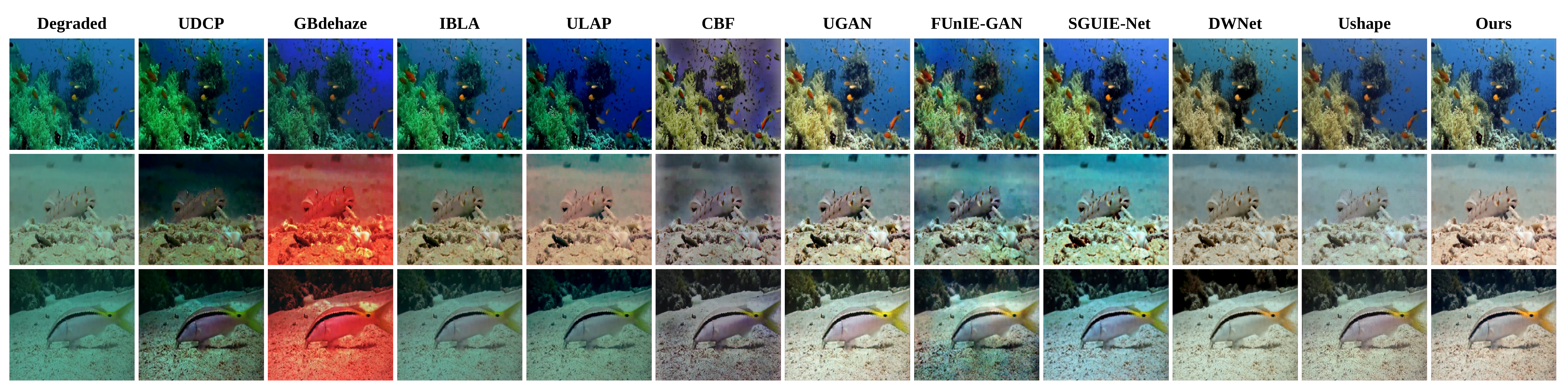}
\caption{Visual demonstration against the state-of-the-art UIE models on UIEB challenge test set.}
\label{uieb_challeng}
\end{figure}

In this subsection, we have presented the qualitative performance of the Lit-Net in both the UIE and UISR tasks. Also, we have carried out the qualitative evaluation with the other existing works. The visual comparison of UIE is shown in Fig. \ref{uieb}, \ref{suim_e}, \ref{euvp}, and \ref{uieb_challeng}. The results displayed in these figures indicate that Lit-Net generates the most visually enhanced images with respect to ground truth. From Fig. \ref{uieb}, one can easily notice that the existing methods IBLA \cite{peng2017underwater} and ULAP \cite{song2018rapid} fail to recover degraded images. The methods CBF~\cite{ancuti2017color} exhibit issues of color deviation in their enhanced images. UGAN \cite{fabbri2018enhancing} and FUnIE-GAN \cite{islam2020fast} suffer from low contrast. SGUIE-Net \cite{qi2022sguie} provides a better result, but it has a bluishness in the produced image. It can be observed from the third row of Fig. \ref{suim_e} that images produced by the proposed model look more perceptually improved and close to the ground truth. This may be due to the fact that the proposed approach better balances visual quality and color correction. It is also observed from Fig. \ref{euvp} that the proposed Lit-Net model is capable of handling enhancement effectively, unlike other methods that often result in under and over-saturation, artifacts or fail to correct color deviations. Similar results on the test set of the UIEB challenge are presented in Fig. \ref{uieb_challeng} to show the efficacy of our approach. Our proposed method produces more consistent enhancement over the state-of-art schemes in terms of color correction, contrast enhancement, and detail improvement.

\begin{figure}[ht]
\centering
\includegraphics[width=4.8in, height = 1.4in]{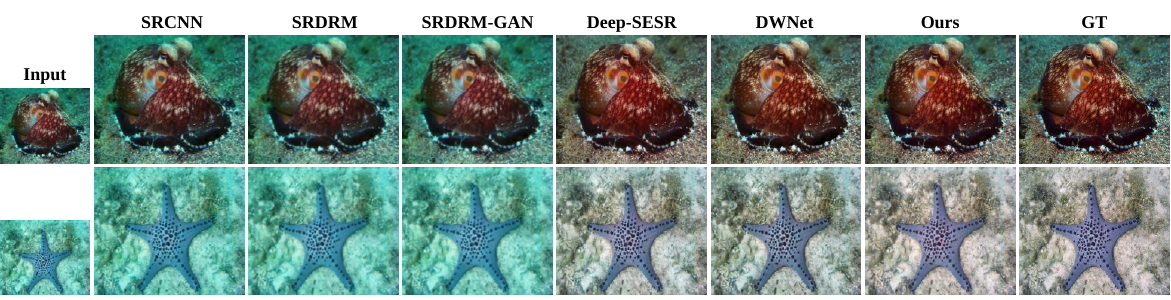}
\caption{Visual demonstration against the state-of-the-arts UISR model on UFO-120 dataset for scale 2.}
\label{sr_2x}
\end{figure}

\begin{figure}[ht]
\centering
\includegraphics[width=4.8in ,height = 1.4in]{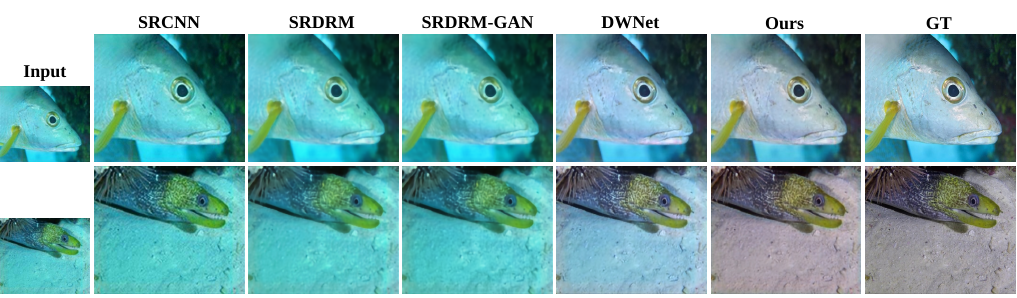}
\caption{Visual demonstration against the state-of-the-arts UISR model on UFO-120 dataset for scale 4.}
\label{sr_4x}
\end{figure}

Fig. \ref{sr_2x} and \ref{sr_4x} depict the qualitative result for the UISR task. It is observed that SRDRM and SRDRM-GAN \cite{islam2020underwater} can not completely eliminate the color deviations in resulting images. Further, the results achieved using Deep SESR \cite{islam2020simultaneous} and DWNet \cite{sharma2023wavelength} still have original noise traces. Fig. \ref{sr_2x} and \ref{sr_4x} show that Lit-Net produces super-resolution (SR) images with finer texture details and fewer distortions than other recent methods. This demonstrates that Lit-Net is more effective in reducing noise and enhancing image quality. Intuitively, the proposed method has the ability to recreate high-frequency details with sharp edges and fewer unwanted distortions compared to previous methods.
\subsection{\textbf{Ablation Study}}
In this section, we analyze the contributions of various parameters in the proposed method. 
\begin{figure}[ht]
\centering
\includegraphics[width=\textwidth , height = 1.8in]{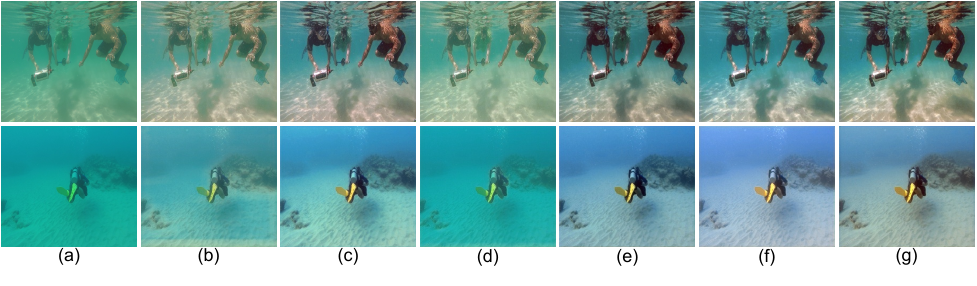}
\caption{Visual demonstration of different modules of Lit-Net. (a) Degraded image, (b) MRAN without attention module, (c) MSAN without spatial attention module, (d) Lit-Net without attention module, (e) Lit-Net with fixed size kernel, (f) Lit-Net without channel-wise split, and (g) Full Lit-Net}
\label{ablation}
\end{figure}
\begin{table*}[ht]
  \caption{Effectiveness of various cost functions on three datasets: EUVP, UIEB, and SUIM-E, for the task of UIE}
  \label{tab:loss_en}
  \centering
  \setlength{\tabcolsep}{1pt}
  \tiny
  \begin{tabular}{c|ccc|ccc|ccc}
    \hline
    &&EUVP&&&UIEB&&&SUIM-E&\\
    \hline
    &$cl_{1}$&$cl_{1}+\lambda_{p}l_{p}$&$cl_{1}+\lambda_{p}l_{p}+\lambda_{s}l_{s}$&$cl_{1}$&$cl_{1}+\lambda_{p}l_{p}$&$cl_{1}+\lambda_{p}l_{p}+\lambda_{s}l_{s}$&$cl_{1}$&$cl_{1}+\lambda_{p}l_{p}$&$cl_{1}+\lambda_{p}l_{p}+\lambda_{s}l_{s}$\\
    \hline
    PSNR&29.340&29.095&\textcolor{red}{29.477}&22.97&23.593&\textcolor{red}{23.603}&24.721&24.899&\textcolor{red}{25.326}\\
    SSIM&0.830&0.849&\textcolor{red}{0.851}&0.849&0.859&\textcolor{red}{0.863}&0.874&0.881&\textcolor{red}{0.884}\\
    UIQM&2.980&3.008&\textcolor{red}{3.027}&3.118&3.010&\textcolor{red}{3.145}&2.819&2.885&\textcolor{red}{2.938}\\
  \hline
\end{tabular}
\end{table*}
\subsubsection{\textbf{Impact of loss functions}}
We have performed a series of experiments to demonstrate the influence of various loss functions incorporated in our proposed model. From Table \ref{tab:loss_en}, it is evident that $l_{p}$ loss, when added with $cl_{1}$ loss, has improved the UIE performance. The intuitive reason behind such improvement may be the low-level features in the improved images may be benefited from the perceptual loss. Moreover, it is observed from Table that incorporating the SSIM-based loss function ($l_{s}$) has improved the corresponding PSNR and SSIM values noticeably. In the case of underwater SISR, Table \ref{tab:loss_sr} shows that the $cl_{1}$ loss produces the least amount of performance. Even though the addition of perceptual loss $l_{p}$ over $cl_{1}$ gained performance in a few resolution configurations, there has been a slight improvement, but not in all metrics. Adding $l_{s}$ loss on top of the earlier losses has improved the performance for all the resolution configurations.
\begin{table}[ht]
  \caption{Effectiveness of various loss functions on UFO-120 dataset, for the task of UISR}
  \label{tab:loss_sr}
  \centering
  \small
  \setlength{\tabcolsep}{3pt}
  \begin{tabular}{c|ccc|ccc|ccl}
    \hline
    Methods &  & PSNR & &  & SSIM & &  & UIQM &\\
    \hline
     & 2x & 3x & 4x & 2x & 3x & 4x & 2x & 3x & 4x\\
    \hline
    $cl_{1}$&23.56&24.97&25.22&.81&.78&.77&2.81&2.90&2.81\\
    $cl_{1}+\lambda_{p}l_{p}$&24.62&25.02&24.96&.79&.78&.79&2.87&2.96&2.88\\
    $cl_{1}+\lambda_{p}l_{p}+\lambda_{s}l_{s}$&\textcolor{red}{25.78}&\textcolor{red}{25.30}&\textcolor{red}{25.32}&\textcolor{red}{.82}&\textcolor{red}{.80}&\textcolor{red}{.80}&\textcolor{red}{3.00}&\textcolor{red}{2.99}&\textcolor{red}{2.93}\\
    \hline
  \end{tabular}
\end{table}
\begin{table*}[ht]
  \caption{Effectiveness of various module on three datasets: UIEB, SUIM-E, and UFO-120(4x) for the task of UIE and UISR}
  \label{tab:ablation}
  \centering
  \setlength{\tabcolsep}{1pt}
  \tiny
  \begin{tabular}{l|cc|ccc|ccc|ccl}
    \hline
    Ablations&&&&UIEB&&&SUIM-E&&&UFO-120\\
    \hline
    &GFLOPS&Parameter(M)&PSNR&SSIM&UIQM&PSNR&SSIM&UIQM&PSNR&SSIM&UIQM\\
    \hline
    MRAN w/o-Attention&17.852&0.5433&22.958&0.859&3.110&22.230&0.862&2.798&25.137&0.800&2.913\\
    MSAN w/o-Attention&17.863&0.5443&23.255&0.860&3.113&24.499&0.884&2.856&25.156&0.798&2.900\\
    Lit-Net w/o-Attention&17.837&0.5373&20.576&0.832&3.124&21.821&0.854&2.738&25.055&0.796&2.922\\
    Lit-Net w fixed-size kernel&15.912&0.3199&23.568&0.855&3.042&24.871&0.882&2.796&24.814&0.795&2.914\\
    Lit-Net w/o-channel split&14.587&0.4917&22.517&0.854&3.073&24.622&0.872&2.833&24.899&0.803&2.877\\
    Lit-Net w L1 loss&17.871&0.5446&23.530&0.857&3.069&25.003&0.877&2.791&25.066&0.801&2.882\\
  \hline  Lit-Net&17.871&0.5446&\textcolor{red}{23.603}&\textcolor{red}{0.863}&\textcolor{red}{3.145}&\textcolor{red}{25.117}&\textcolor{red}{0.884}&\textcolor{red}{2.918}&\textcolor{red}{25.326}&\textcolor{red}{0.803}&\textcolor{red}{2.938}\\
  \hline
\end{tabular}
\end{table*}
\subsubsection{\textbf{Effect of Attention Module}}
In Table \ref{tab:ablation}, we have shown the impact of attention modules in our Lit-Net. We have carried out the experiments on all the datasets. Here, we have presented the result of the UIEB, SUIM-E, and UFO-120 (4x) test sets. From the first, second, and third rows of Table \ref{tab:ablation}, it is observed that the performance of the proposed scheme is relatively low when the attention module is not included in the MRAN block, MSAN block, and the whole Lit-Net. Intuitively, the addition of the CBAM module helps the proposed Lit-Net model to effectively preserve and recover the texture details of the degraded images. As shown in Fig. \ref{ablation}(b), (c), and (d), the enhanced image without the attention module fails to correct the color, local and global contrast.

\subsubsection{\textbf{Effect of different receptive fields and encoder block}}
To show the effect of receptive fields and encoder block, we have carried out the experiments with the following setup:
\begin{enumerate}
	\item We have set  $3\times3$ kernel size in the MRAN instead of three different kernels.
	\item We have set one $1\times1$ kernel in the encoder of MSAN instead of four parallel $1\times1$.
\end{enumerate}

The fourth row of Table \ref{tab:ablation} shows a substantial drop in SSIM, PSNR, and UIQM when we have performed the task mentioned above. Notably, Lit-Net with this setting has decreased the $0.15\%$ and $0.98\%$ PSNR value and $3.28\%$ and $4.18\%$ UIQM value on UIEB and SUIM-E datasets, respectively. Fig. \ref{ablation}(e) shows the visual result of the Lit-Net with these settings. It can be seen that the produced result fails to recover the color.
\subsubsection{\textbf{Effect of weighted color specific L1 loss and color channel split module}}
The weighted color specific $l_{1}$ loss is designed to correct the color, which is affected by the wavelength of the underwater medium. The sixth row of Table \ref{tab:ablation} shows the result of our method with $l_{1}$ loss along with other losses. It is observed that the proposed method with $l_{1}$ loss has decreased the PSNR, SSIM, and UIQM values as compared to $cl_{1}$ loss.

Similar to color-specific loss, Lit-Net processes the R, G, and B channels with different kernel sizes to capture the global context from the blue channel and the local context from the red channel. When the proposed method processes the entire RGB image instead of employing R, G, and B channel-wise processing, it can be observed from the $5th$ row of Table \ref{tab:ablation}, Lit-Net provides a decrease of $1.086$ dB, $0.495$ dB, and $0.427$ dB PSNR. Compared with Lit-Net w/o-channel split, the full model improves the UIQM score with $2.29\%$, $2.91\%$, and $2.08\%$ on UIEB, SUIM-E, and UFO-120 (4X) test set, respectively. Fig. \ref{ablation}(f) shows the visual quality of Lit-Net w/o-channel split, it is less sharp and fails to recover the details. Fig. \ref{ablation}(g) shows that the full Lit-Net provides better color correction, detail recovery, and perceptual enhancement.
\subsection{\textbf{Impact on High level Vision Applications}}
We have demonstrated the resilience of the improved images generated by the Lit-Net against state-of-the-art UIR strategies on several underwater high-level vision applications. We have taken into account tasks like underwater object detection and single-image semantic segmentation for this.
\subsubsection{\textbf{Underwater Semantic Segmentation}}
\begin{figure}[ht]
\centering
\includegraphics[width= 5in, height = 2.5in]{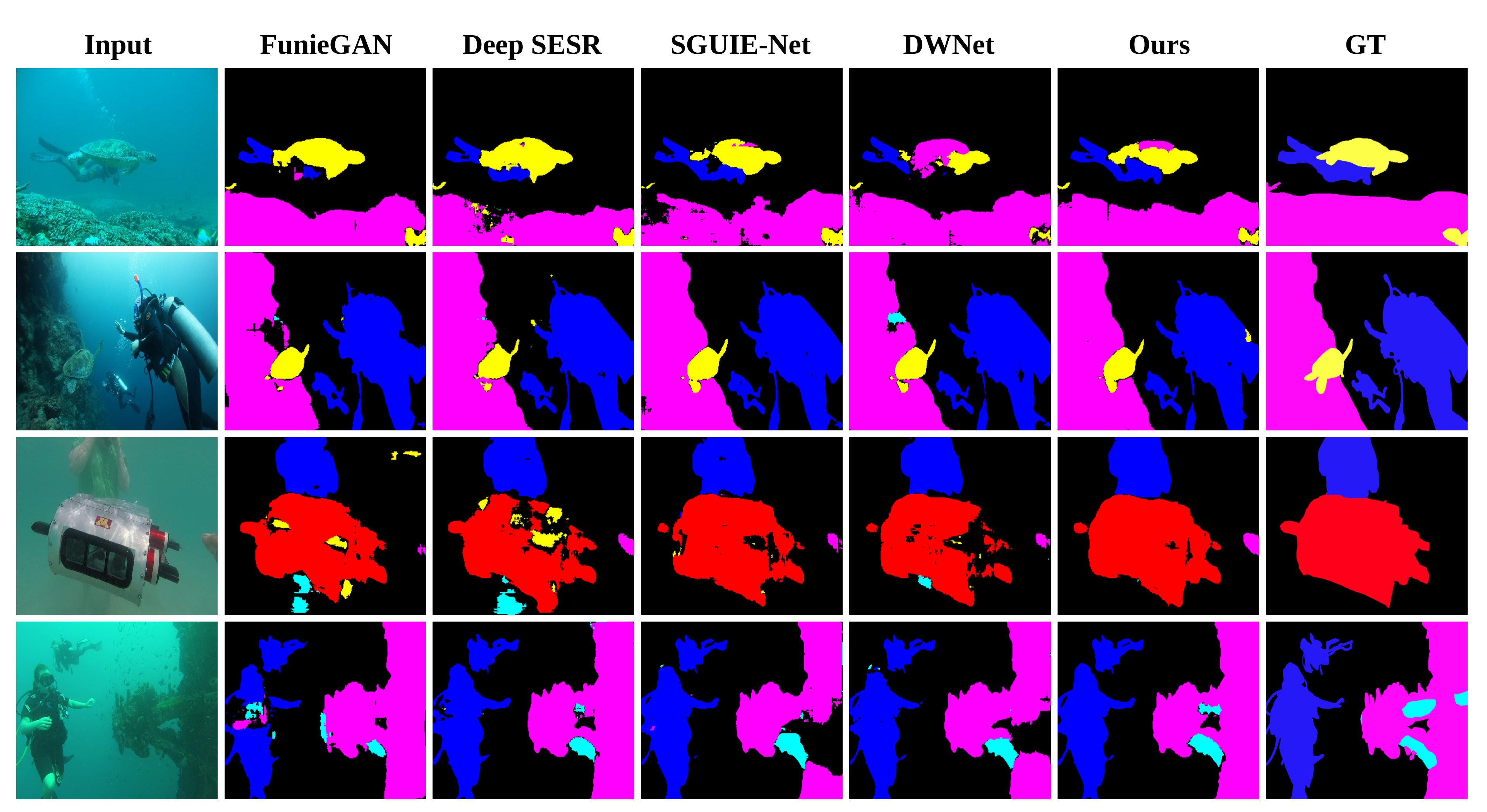}
\caption{Visual comparison of the semantic segmentation maps derived from the improved underwater images from four existing UIE works.}
\label{suim}
\end{figure}
Semantic segmentation is a process of labelling or assigning a category to each pixel in an image. This technique is utilized to identify a group of pixels that collectively belong to separate categories for detailed scene understanding. We have used the SUIM dataset \cite{islam2020suim} for underwater semantic segmentation. First, we enhance the images in the SUIM dataset using our proposed Lit-Net. Second, we utilized the SUIM-Net \cite{islam2020suim} method for the semantic segmentation task to show the performance gain by using the enhanced images. Fig. \ref{suim} shows that generated segmentation maps are more refined than existing ones. 
\begin{table}[ht]
  \caption{Mean average precision (mAP) comparison on detection task}
  \label{tab:detection}
  \centering
  \small
  \setlength{\tabcolsep}{2.5pt}
  \begin{tabular}{ccccccccl}
    \hline
     &Fish &Jellyfish &Penguin &Puffin &Shark &Starfish &Stingray &All\\
    \hline \hline
    Degraded&0.815&0.929&0.749&\textcolor{blue}{0.602}&0.749&\textcolor{blue}{0.838}&0.781&0.780\\
    FUNIE-GAN &0.744 &0.923 &0.654 &0.445 &0.664 &0.716 &0.755 &0.700\\
    Deep SESR &0.706 &\textcolor{blue}{0.943} &0.643 &0.41 &0.638 &0.703 &0.757 &0.686\\
     DWNet &\textcolor{red}{0.839} &\textcolor{red}{0.948} &\textcolor{blue}{0.752} &0.551 &\textcolor{blue}{0.757} &0.819 &\textcolor{blue}{0.806} &\textcolor{blue}{0.782}\\
    Lit-Net &\textcolor{blue}{0.821} &0.940 &\textcolor{red}{0.753} &\textcolor{red}{0.615} &\textcolor{red}{0.785} &\textcolor{red}{0.841} &\textcolor{red}{0.814} &\textcolor{red}{0.796}\\
     \hline
\end{tabular}
\end{table}
\begin{figure}[ht]
\centering
\includegraphics[width=3.6in, height = 2.4in]{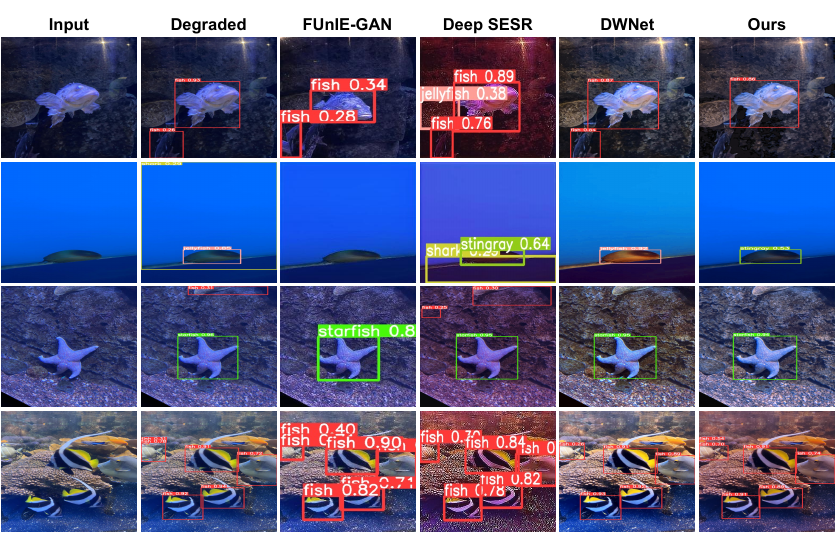}
\caption{Visual demonstration of object detection by utilizing the improved images from three existing UIE works.}
\label{detect}
\end{figure}
\subsubsection{\textbf{Underwater Object Detection}}
To validate the impact of UIE on object detection tasks, we utilize the improved outcomes in a detection algorithm. For this task, we have used the YOLOv5 \cite{yolov5} and Aquarium dataset \cite{aquarium}. We have reported the quantitative results in Table \ref{tab:detection}. The results indicate that the proposed method exhibits better detection accuracy than the other competitive methods. Fig. \ref{detect} displays the visualization of detection results obtained on the Aquarium dataset. It is shown in Fig. \ref{detect} that in spite of visual enhancement, FUNIEGAN \cite{islam2020fast}, Deep SESR \cite{islam2020simultaneous}, and DWNet \cite{sharma2023wavelength} methods showed inferior results for the detection performance in comparison with proposed Lit-Net model. 
\begin{figure}[ht]
\centering
\includegraphics[width=3in, height = 2in]{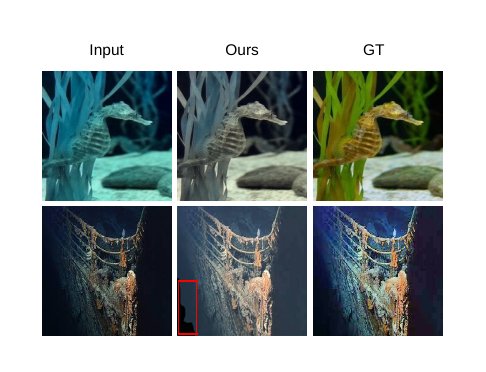}
\caption{Visual demonstration of object detection by utilizing the improved images from three existing UIE works.}
\label{fail}
\end{figure}
\subsection{\textbf{Failure Case}}
We examine a few failure cases of Lit-Net, as depicted in Figure \ref{fail}. Our approach needs to perform well in a few cases where the rate degradation is highly intense, such as extremely low contrast, irregular coloring, etc. For example, in the first image, the GT image contains color for the object and sea grass, and it is almost difficult to observe its color in the predicted image because it is tough to identify the colors from the original degraded image. As illustrated in the second image, the contrast of the lower left region is extremely diverse from the contrast of the lower region. Since our model lacks explicit provisions for extremely low-light image enhancement, a few resultant images may exhibit a tendency towards suboptimal enhancement. In the future, we can introduce extremely low-light correction modules to precisely enhance images with uneven illumination.
\section{Conclusion}\label{secC}
In this paper, a novel light-weight image enhancement and super resolution model, called Lit-Net, is proposed specifically for underwater images. In this model, there are two main modules, namely Multi-resolution Attention Network (MRAN) and Multi-scale Attention Network (MSAN). Different receptive fields (different convolution kernels) have been used to achieve the multi-resolution image analysis while retaining the original input resolution. In the MSAN module, we have used spatial attention based skip connections between respective encoder-decoder layers to get both semantically and spatially rich feature representations. We also demonstrated that employing this type of encoder design aids in learning diverse features and reduces the complexity of the model. Furthermore, we used the weighted color channel specific $l_{1}$ loss which enhances the performance of Lit-Net. Experiments on both UIE and UISR show that the proposed Lit-Net outperforms the state-of the art methods. We also provided an ablation studies to justify the contribution of different modules. It is also experimentally demonstrated that the images, enhanced with the proposed model, achieve significant improvements in the performance of various advanced visual tasks, such as underwater object detection and semantic segmentation.

\bibliography{sn-article}

\end{document}